\documentclass{article}
\usepackage{ijcai18}

\usepackage{times}
\usepackage{xcolor}
\usepackage{soul}
\usepackage[utf8]{inputenc}
\usepackage[small]{caption}
\usepackage{algorithm}
\usepackage{algorithmic}
\usepackage{amssymb}
\usepackage{setspace}
\usepackage{graphics}
\usepackage{multirow}

\title{Automatic Construction of Parallel Portfolios via Explicit Instance Grouping}

 \author{
Shengcai Liu,
$^1$, 
Ke Tang,
$^2$, 
Xin Yao
$^2$, 
\\
$^1$ USTC-Birmingham Joint Research Institute in Intelligent Computation
and Its Applications (UBRI),\\ 
School of Computer Science and Technology,
University of Science and Technology of China, Hefei, Anhui, China \\
$^2$ Department of Computer Science and Engineering, Southern University of Science and Technology, Shenzhen, China\\
liuscyyf@mail.ustc.edu.cn,
tangk3@sustc.edu.cn,
xiny@sustc.edu.cn
}

\begin{document}

\maketitle

\begin{abstract}
Simultaneously utilizing several 
complementary solvers is a simple yet
effective strategy for solving 
computationally hard problems. 
However, manually building 
such solver portfolios typically
requires considerable domain knowledge and plenty 
of human effort. 
As an alternative, automatic construction of 
parallel portfolios (ACPP) aims at automatically 
building effective parallel portfolios based on a 
given problem instance set and a given rich design space. 
One promising way to solve the 
ACPP problem is to explicitly group 
the instances into different subsets and 
promote a component solver to handle each of them. 
This paper investigates solving ACPP from this 
perspective, and especially studies 
how to obtain a good instance grouping.
The experimental results showed 
that the parallel portfolios constructed by 
the proposed method could achieve consistently
superior performances to the ones 
constructed by the state-of-the-art ACPP methods, 
and could even rival sophisticated
hand-designed parallel solvers.   
\end{abstract}

\section{Introduction}

It has been widely observed in many problem domains 
\cite{xu2010hydra,tang2014population}
that there is no universal optimal solver dominating all other solvers on all problem instances. Instead, different solvers perform well on different problem instances.
Thus a natural idea is to combine those complementary solvers together to achieve a better overall performance. Typical examples include algorithm selection methods \cite{rice1976algorithm,xu2008satzilla,kotthoff2016algorithm} 
which try to select the best solver for every single problem instance before solving it, and adaptive solvers such as adaptive parameter control 
\cite{karafotias2015parameter},
reactive search \cite{battiti2008reactive} and 
hyper-heuristics \cite{burke2013hyper} 
which seek to dynamically determine the best solver setting while solving a problem instance. 
In principle, all these methods need to involve some mechanisms (e.g., selection or scheduling) to appropriately allocate computational resource to different solvers.

Recently parallel portfolios \cite{gomes2001algorithm,tang2014population,lindauer2017automatic} as another paradigm of simultaneously utilizing several sequential solvers, have attracted more and more research interest. While solving a problem instance, parallel portfolios run all the component solvers in parallel until the first of them solves it; thus the performance of a parallel portfolio 
is always the best performance achieved by its component solvers. 
Different from algorithm selection and adaptive solvers, parallel portfolios do not necessarily require any extra resource allocation since  each involved component solver is simply assigned with the same amount of resource. Moreover, the rapid growth of parallelism in computational power \cite{gepner2006multi} makes such parallel solving strategy more and more critical for solving computationally hard problems.
However, the manual construction of parallel portfolios is non-trivial. Specifically, identifying (or designing) a set of relatively uncorrelated sequential solvers which are complementary to each other still requires relatively significant domain knowledge \cite{xu2010hydra}. 

Recently \cite{lindauer2017automatic} proposed using automatic construction of parallel portfolios (ACPP) as a first step towards tackling parallel portfolio construction. The goal of ACPP is to automatically construct a parallel portfolio based on the rich design space induced by a highly parametrized sequential solver or a set of them. More formally, let $C$ and $I$ denote the configuration space 
of the parameterized solvers
and the given set of problem instances, respectively. The parallel portfolio 
with $k$ component solvers could be denoted as a $k$-tuple, i.e., 
$P=(c_{1},...,c_{k})$, in which $c_{i} \in C$ is an individual configuration and represents the $i$-th component solver of $P$. The goal is to
find $c_{1},...,c_{k}$ from $C$, such that the performance of $P$ on $I$ according to a given metric $m$ is optimized.

There are three key ACPP methods, namely GLOBAL, $\mathrm{PARHYDRA}$ and CLUSTERING, in which GLOBAL and 
$\mathrm{PARHYDRA}$ are both proposed by \cite{lindauer2017automatic} while CLUSTERING is adapted 
by \cite{lindauer2017automatic} from $\mathrm{ISAC}$ \cite{kadioglu2010isac} for comparison \footnote{Although $\mathrm{ISAC}$ is an automatic portfolio construction method for algorithm selection, 
its basic idea is also applicable to ACPP. See \cite{lindauer2017automatic} for more details.}.
Generally, these methods can be divided into two categories. GLOBAL belongs to the first category, which considers ACPP an algorithm configuration problem by directly treating $P$ as a parameterized solver. By this means powerful automatic algorithm configurators such as ParamILS \cite{hutter2009paramils}, GGA \cite{ansotegui2009gender}, irace \cite{ansotegui2009gender} and SMAC \cite{hutter2011sequential} could be directly applied to configure $P$ (GLOBAL uses SMAC in \cite{lindauer2017automatic}). The key issue of GLOBAL is that its scalability is limited since the size of the configuration space of $P$, i.e., $|C|^{k}$, increases exponentially with the number of the component solvers, i.e., $k$.
The other two methods, $\mathrm{PARHYDRA}$ and CLUSTERING, solve the ACPP problem from the perspective of
instance grouping. That is, they explicitly or implicitly promote different component solvers of $P$ to handle different subsets of problem instances, with the goal that the resulting component solvers would be complementary to each other.
More specifically, starting from an empty portfolio, $\mathrm{PARHYDRA}$ proceeds iteratively and in the $i$-th iteration it uses an algorithm configurator (also SMAC) to configure $c_{i}$ to add to the current portfolio, i.e., $(c_{1},...,c_{i-1})$, such that the performance of the resulting portfolio, i.e., $(c_{1},...,c_{i-1},c_{i})$, is optimized.
In other words, in each iteration $\mathrm{PARHYDRA}$ intrinsically aims to find a solver 
that can improve the current portfolio to the best extent.
Since the greatest chance of the current portfolio getting improved is on those problem instances which cannot be solved satisfactorily, thus while $\mathrm{PARHYDRA}$ configuring the $i$-th component solver $c_{i}$, the configurator would actually promote $c_{i}$ to handle those intractable instances to the current portfolio $(c_{1},...,c_{i-1})$.
Compared to $\mathrm{PARHYDRA}$, CLUSTERING  
adopts a more explicit way to group instances. It clusters the problem instances in a given (normalized) instance feature space and then independently configures (using SMAC) a component solver on each instance cluster.
The advantage of PARHDYRA and CLUSTERING is that they keep the size of the configuration space
involved in each algorithm configuration 
task as $|C|$.


The main issue of $\mathrm{PARHYDRA}$ is that its 
intrinsic greedy mechanism may cause 
stagnation in local optima.
To alleviate this problem, \cite{lindauer2017automatic} makes a modification to $\mathrm{PARHYDRA}$ by allowing simultaneously configuring several component solvers in each iteration. The resulting method is named 
$\mathrm{PARHYDRA_{b}}$, where $b\ (b \geq 1)$ is  the number of the component solvers configured in each iteration.
$\mathrm{PARHYDRA}$ and GLOBAL could be both seen as special cases of $\mathrm{PARHYDRA_{b}}$ with $b=1$ and $b=k$ respectively.
It is conceivable that the choice of $b$ 
is very important for $\mathrm{PARHYDRA_{b}}$ 
since the tendency to stagnate in local optima
would increase as $b$ gets smaller,
while the size of the configuration space involved in each configuration task in $\mathrm{PARHYDRA_{b}}$,
i.e., $|C|^{b}$, would grow exponentially 
as $b$ gets larger.
However, in general the best value
of $b$ may vary across different scenarios,
and for a specific scenario
it is very hard to determine a good choice of
$b$ in advance.

For methods based on explicit instance grouping such as CLUSTERING, 
obviously the quality of the instance grouping is crucial.
A good instance grouping should meet at least one requirement:
The instances that are grouped together should
be similar in the sense that in the configuration space $C$ 
there exist same good 
configurations for them.
CLUSTERING uses the distances in the normalized feature space
to characterize such
similarity. Although the problem features have been
proved very useful for modeling the relationship between 
problem instances and solvers 
(e.g., algorithm selection), some practical issues still exist 
while applying CLUSTERING to ACPP.
Specifically, the used instance features
as well as the normalization of the features would greatly influence
the final clustering results.
However, determining appropriate 
choices of them is very hard 
since accurate assessment of the 
cluster quality relies on completely constructed portfolios.

In this paper, we investigate further solving 
the ACPP problem based on explicit instance
grouping. Specifically, we propose 
a new ACPP method
named parallel configuration with instance 
transfer (PCIT) \footnote{Here ``parallel configuration'' means
that the configuration processes of the component solvers are 
independent of each other and thus could be conducted in parallel.},
which explicitly divides 
the instances into different subsets and 
configures a component solver on each subset. 
The most novel feature of PCIT is its dynamic 
instance transfer mechanism.
Unlike CLUSTERING which determines the instance grouping in advance and then keeps it 
fixed through the whole process,
during portfolio construction 
PCIT would dynamically adjust 
the instance grouping  
by transferring instances between 
different subsets.
The instance transfer is conducted
with the goal that the instances 
which share the same high-quality 
configurations (in the configuration space $C$)
would be grouped together,
such that the complementarity between the component solvers
configured on different subsets would be favourably enhanced.
The experimental results showed that the parallel portfolios 
constructed by PCIT could achieve consistently
superior performances to the ones output
by the existing ACPP methods, and could 
even rival the state-of-the-art hand-designed parallel solvers. 

\section{Proposed Method}
The basic idea of PCIT is simple. 
Although it is hard to obtain an appropriate 
instance grouping at one stroke, 
it is possible to gradually
improve an instance grouping.
PCIT adopts a random splitting strategy to obtain
an initial grouping; that is, the instances are 
evenly and randomly divided into $k$ subsets. 
It is conceivable that the quality of random instance grouping is 
not guaranteed at all since there is no guidance 
involved in the grouping procedure.
Consider a simple example where instance set 
$I=\{ins_{1}, ins_{2}, ins_{3}, ins_{4}\}$, 
configuration space $C=\{\theta_{1}, \theta_{2}\}$, 
$ins_{1},ins_{2}$ shares 
the high-quality configuration $\theta_{1}$
and $ins_{3},ins_{4}$ shares 
the high-quality configuration $\theta_{2}$.
Obviously the appropriate grouping for 
this example is $\{ins_{1},ins_{2}\} \{ins_{3}, ins_{4}\}$, which 
would lead algorithm configurator to 
output $\theta_{1}$ and $\theta_{2}$ on the first and the second 
subset respectively, thus producing the optimal  
portfolio $P=\{\theta_{1}, \theta_{2}\}$.
Random splitting strategy may fail on this example if it happens to
split $I$ as $\{ins_{1},ins_{3}\} \{ins_{2}, ins_{4}\}$ or
$\{ins_{1},ins_{4}\} \{ins_{2}, ins_{3}\}$,
which could cause algorithm configurator to output the
same component solver i.e., $(\theta_{1}, \theta_{1})$ or 
$(\theta_{2}, \theta_{2})$, on both subsets.


The key point here is that if the problem instances  
grouped together do not share the same high-quality configurations, 
then the cooperation between the component solvers 
configured on these subsets would be much affected,
thus limiting the quality of the final output parallel portfolio.
To handle this issue, PCIT employs an instance transfer 
mechanism to improve the instance grouping
during the construction process
by transferring instances between different subsets.
More specifically, as the configuration process 
of a component solver on a subset proceeds, 
if the algorithm configurator cannot 
manage to find a common 
high-performance configuration for 
every instance in the subset but only some of them, 
then it can be inferred that these intractable 
instances may correspond to different 
high-quality configurations (in the configuration space $C$) from others. 
It is therefore better to transfer these instances 
to other subsets that are more suitable to them.

PCIT conducts the instance transfer with the help of
incumbent configurations (i.e., the best configurations found by the algorithm configurator). 
In each subset, the instances which cannot be solved 
satisfactorily by the corresponding incumbent are 
identified as the ones that need to be transferred, 
and the target subset of each transferred instance is 
determined according to how well the incumbent on
the candidate subset 
could perform on the instance. 
In essence, the incumbent on a subset can be seen as 
a common special characteristic of those ``similar'' instances
(in the sense they share the same high-quality configurations)
within the subset, and PCIT uses it to identify 
those ``dissimilar'' instances and find better 
subsets for them.
In each subset, the performance of the incumbent 
on each instance could be obtained from the rundata 
collected from the configuration process.
However, while determining the target subsets 
for the transferred instances, 
how well the incumbents on the candidate subsets  
would perform on the transferred instances are unknown.
One way to obtain these performances 
is to actually test these incumbents on the transferred instances,
which however would introduce considerable 
additional computational costs.
To avoid this, PCIT builds empirical performance
models (EPM) \cite{hutter2014algorithm}
based on the collected rundata to 
predict these performances.  

\subsection{Algorithm Framework}

The pseudo-code of PCIT is given in Algorithm~\ref{PCIT}.
The main difference between PCIT and the
existing methods (e.g., GLOBAL and CLUSTERING) is that 
in PCIT the portfolio construction process is divided into 
$n$ (we always set $n=4$ in this paper) 
sequential phases (Lines 3-13 in Algorithm~\ref{PCIT}).
The first $(n-1)$ phases serve as adjustment phases,
in each of which the instance grouping 
is adjusted (Line 12) once the configuration 
procedures for all 
component solvers (Lines 9-11) finish. 
The last phase is the construction
phase in which the component solvers
of the final portfolio are configured on the 
obtained subsets with a large amount of time.
In fact, the time consumed for the
configuration processes
in the last phase amounts to the sum of the time 
consumed for the configuration processes in 
the first $(n-1)$ phases (Lines 4-8). 
One thing which is not detailed in Algorithm~\ref{PCIT} 
for brevity is that, on each subset, to keep the continuity 
of the configuration processes across successive phases, 
the incumbent configuration obtained in the 
previous phase is always used to initialize the configuration
procedure in the next phase.

\captionsetup[algorithm]{labelsep=colon}
\begin{algorithm}
\caption{PCIT}
\label{PCIT}
\begin{flushleft}
      \textbf{Input:}  parameterized solvers with configuration space $C$; number of component solvers $k$; instance set $I$; performance metric $m$; configurator $AC$; number of independent runs of portfolio construction $r_{pc}$; time budget for configuration process $t_{c}$; time budget for  validation process $t_{v}$; number of stages $n$; features $F$ for all instances in $I$\\
      \textbf{Output:} parallel portfolio $(c_{1},..,c_{k})$
\end{flushleft}
\begin{algorithmic}[1]
\FOR{$i:=1...r_{pc}$}
\STATE Randomly and evenly split $I$ into $I_{1},...,I_{k}$
\FOR{$phase:=1...n$}
\IF{$phase$ = $n$}
\STATE $t \gets \frac{t_{c}}{2}$
\ELSE
\STATE $t \gets \frac{t_{c}}{2(n-1)}$
\ENDIF
\FOR{$j:=1...k$}
\STATE Obtain component solver $c_{j}$ by running $AC$ on configuration space $C$ on $I_{j}$ using $m$ for time $t$
\ENDFOR
\STATE $I_{1},...I_{k} \gets \textrm{InsTransfer}(I_{1},...I_{k}, c_{1},...c_{k}, F)$ 
\ENDFOR
\STATE $P_{i} \gets (c_{1},...,c_{k})$ 
\ENDFOR
\STATE Validate each of $P_{1},...,P_{r_{pc}}$ on $I$ using $m$ for time $t_{v}$
\STATE Let $P$ be the portfolio which achieved the best validation performance
\RETURN{$P$}
\end{algorithmic}
\end{algorithm}

Another important difference between PCIT and the existing methods 
lies in the way of obtaining reliable outputs. 
For existing methods, the uncertainty of the portfolio 
construction results
mainly comes from the randomness of the output of
the algorithm configurator (especially when the given parametrized solvers 
are not deterministic).
Thus for each specific algorithm configuration task, 
typically they conduct multiple independent runs of the 
algorithm configurator 
(with different random seeds),
and then validate the configurations produced by these runs 
to determine the output one. 
For PCIT, in addition to the randomness mentioned above,
a greater source of uncertainty is
the randomness of the initial instance grouping results.
One way to handle both of them is to perform
multiple runs of portfolio construction 
(with different initial instance groupings),
and in each construction process the algorithm configurator is also run for multiple times for each configuration task.
In this paper, to keep the design simple, we only 
allow repeated runs of portfolio construction and rely on the 
validation to ensure the reliability 
of the final output (Lines 16-18 in Algorithm~\ref{PCIT}).

PCIT can be easily performed in parallel.
First, different portfolio construction 
runs (Lines 1-15 in Algorithm~\ref{PCIT}) can be executed
in parallel, and second, during each construction run
the configuration processes for different component 
solvers (Lines 9-11 in Algorithm~\ref{PCIT}) can also 
be executed in parallel.

\subsection{Instance Transfer}

As shown in Algorithm~\ref{IT}, the instance transfer 
procedure first builds an empirical performance model (EPM) 
based on the rundata collected from all the 
previous algorithm configuration processes (Line 1).
More specifically, the rundata is actually records of 
runs of different solver configurations 
on different instances, and each run 
can be represented by a 3-tuple, i.e., 
$(config, ins, result)$.
The exact implementation of EPM here is the 
same as the one in SMAC \cite{hutter2011sequential},
which is a random forest that takes as input
a solver configuration $config$ and a problem instance $ins$ 
(represented by a feature vector), and predicts 
performance of $config$ on $ins$.
The performances of the incumbent configuration on instances 
in each subset are obtained by querying 
the corresponding runs in rundata 
\footnote{The average performance is used if there
are several such runs. In case that there is no such run in rundata, which means through the configuration process the incumbent configuration has not been tested on the instance yet, the instance 
will be excluded from the whole transfer process.}
(Line 2).
After collecting all of them, 
the median value is used to  
identify the instances which will be transferred
(without loss of generality, we assume a smaller value is better for $m$)
(Line 3).
Then these instances are examined one by one in 
a random order (Lines 7-22),
for each examined instance the target subset is
determined according to three rules (Line 13):
1) 
Both the source subset and the target subset 
will not violate the constraints on the 
subset size
after the instance is transferred;
2)
The predicted performance on the instance is 
not worse on the target subset; 
3) The target subset is the one with the best
predicted performance among the ones 
satisfying 1) and 2).  
The subset size constraints, i.e., 
the lower bound $L$ and the upper bound $U$
in Algorithm~\ref{IT},
are set to prevent the occurrence of too large 
or too small subsets.
In this paper
$L$ and $U$ are 
set as $\lceil(1\pm0.2)\frac{|I|}{k}\rceil$, respectively.
Since the sizes of the subsets keep changing
during the instance transfer process,  
there is a possibility
that an instance, which was examined earlier and 
at that time no target subset satisfying the above
conditions was found, has a satisfactory target 
subset later.
To handle this situation,
instances which are not successfully transferred  
will be examined again in the next round (Line 26),
and the whole procedure will be terminated (Lines 23-25) if
there is no instance which needs to be transferred,
or there is no successful transfer in a round (Lines 7-22). 

\renewcommand\topfraction{0.85}
\renewcommand\bottomfraction{0.85}
\renewcommand\textfraction{0.1}
\renewcommand\floatpagefraction{0.85}
\captionsetup[algorithm]{labelsep=colon}
\begin{algorithm}
\caption{InsTransfer}
\label{IT}
$R$ is the run data collected from 
all the previous algorithm configuration processes.
$L$ and $U$ are the lower bound and the upper bound
of the size of a subset, respectively. 
\begin{flushleft}
      \textbf{Input:} instance subsets $I_{1},...,I_{k}$, 
      incumbent configurations $c_{1},...,c_{k}$, 
      instance features $F$
       \\
      \textbf{Output:} instance subsets $I_{1},...,I_{k}$
\end{flushleft}
\begin{algorithmic}[1]
\STATE Build an $EPM$ based on $R$ and $F$
\STATE For each instance $ins$ in each subset, obtain the performance
of the corresponding incumbent configuration on it from $R$ , denoted as $P(ins)$
\STATE Let $v$ be the median value of all $P(ins)$ across all subsets, and
the instances with bigger values than $v$ are identified as the ones which 
need to be transferred, denoted as $T$
\WHILE{$true$}
\STATE $T_{success} \gets \varnothing$
\STATE $T_{remaining} \gets \varnothing$
\WHILE{$T \ne \varnothing$}
\STATE Randomly select an instance $ins$ from $T$ and 
let $I_{s}$ and $c_{s}$ be the subset containing $ins$
and the corresponding incumbent configuration, respectively
\STATE $T \gets T-\{ins\}$ 
\STATE For each incumbent configuration $c$ of $c_{1},...c_{k}$, 
use $EPM$ to obtain the predicted performance of $c$ on $ins$,
denoted as $E(c)$.
\STATE Sort $c_{1},...c_{k}$ according to 
the goodness of $E(c_{1}),...,E(c_{k})$,
denoted as $c_{\pi(1)},...,c_{\pi(k)}$
\FOR{$j:=1...k$}
\IF{$E(c_{\pi(j)}) \leq E(c_{s}) \ \&\&\ |I_{\pi(j)}| < U \ \&\&\ |I_{s}| > L$}
\STATE $I_{s} \gets I_{s} - \{ins\}, I_{\pi(j)} \gets I_{\pi(j)} \cup \{ins\}$
\STATE $T_{success} \gets T_{success} \cup \{ins\}$
\STATE \textbf{break}
\ENDIF
\ENDFOR
\IF{$ins \notin T_{success}$}
\STATE $T_{remaining} \gets T_{remaining} \cup \{ins\}$
\ENDIF
\ENDWHILE
\IF{$T_{success} = \varnothing \ || \ T_{remaining} = \varnothing$}
\STATE \textbf{break}
\ENDIF
\STATE $T \gets T_{remaining}$
\ENDWHILE
\RETURN{$I_{1},...I_{k}$}
\end{algorithmic}
\end{algorithm}

\subsection{Computational Costs}
\label{runtime}
The computational costs of ACPP methods are mainly composed of
two parts: the costs of configuration processes 
and the costs of validation.
For PCIT, the total CPU time consumed 
is $r_{pc}\cdot k \cdot (t_{c}+t_{v})$ (the small overhead
introduced by instance transfer in PCIT is ignored here).
Similarly, for GLOBAL and CLUSTERING, it is 
$r_{ac}\cdot k \cdot (t_{c}+t_{v})$,
where $r_{ac}$ is the number of independent runs 
of algorithm configurator (for each configuration task).
For $\mathrm{PARHYDRA_{b}}$, the consumed CPU time 
is 
$r_{ac} \cdot \sum_{i=1}^{\frac{k}{b}}[i \cdot b \cdot (t_{c}^{b}+t_{v}^{b})]$,
where $t_{c}^{b}$ and $t_{v}^{b}$ refer in particular to the configuration time budget and the validation time budget used in $\mathrm{PARHYDRA_{b}}$ (See \cite{lindauer2017automatic} 
for more details).


\section{Empirical Study}
We conducted experiments on two widely studied 
domains, SAT and TSP. 
Specifically, we used our method to build 
parallel portfolios based on a training set,
and then compared them with the ones constructed by 
the existing methods,
on an unseen test set.

\subsection{Experimental Setup}
\subsubsection{Portfolio Size and Performance Metric}
We set the number of component solvers to 8 
(same as \cite{lindauer2017automatic}),
since 8-core (and 8-thread) machines are 
widely available now.
The optimization goal considered here is to 
minimize the time required by a solver
to solve the problem (for SAT) or to find 
the optimum of the problem (for TSP).
In particular, we set the performance metric
to Penalized Average Runtime–10 (PAR-10) \cite{hutter2009paramils},
which counts each timeout as 10 times the given
cutoff time.
The optimal solutions for TSP instances 
were obtained using Concorde \cite{applegate2006concorde},
an exact TSP solver.

\subsubsection{Scenarios}
For each problem domain we considered 
constructing portfolios based on a 
single parameterized solver 
and based on multiple 
parameterized solvers, 
resulting in four different scenarios.
For brevity, we use SAT/TSP-Single/Multi 
to denote these scenarios.
Table~\ref{tab:scenario} 
summarizes the used instance sets, 
cutoff time, and base parameterized
solvers in each scenario.
Except in SAT-Multi we reused the
settings from \cite{lindauer2017automatic},
in the other three scenarios 
we all used new settings which had never been
considered before in the literature of ACPP.
We especially note that this was the first
time the ACPP methods were applied to TSP.
Settings in SAT-Multi are 
the same as the ones 
in \cite{lindauer2017automatic}: 
1) Instance set obtained from
the application track of the SAT'12 Challenge
were randomly and evenly split into a training
set and a test set, and to
ensure the computational costs 
for portfolio construction
would not be prohibitively large, 
the cutoff time used in training (180s) 
was smaller 
than the one used in testing 
(900s, same as the SAT'12 challenge);
2) The parameterized solvers in 
SAT-Multi (the configuration space $C$ contains 
150 parameters in total, including a top-level parameter used to select a base solver) 
were the 8 sequential solvers considered by
\cite{wotzlaw2012pfoliouzk} 
when designing pfolioUZK, 
the gold medal winning solver in the 
parallel track of the SAT'12 Challenge. 
In SAT-Single, we chose instances from
the benchmark used in the agile track of 
the SAT'16 Competition for its moderate
cutoff time (60s).
Specifically, we randomly selected 
2000 instances from the original 
benchmark (containing 5000 instances) 
and divided them evenly 
for training and testing.
We chose Riss6 \cite{manthey2016riss},
the gold medal winning solver of this track,
as the base solver.
Since Riss6 exposes
a large number of parameters,  
we selected 135 parameters from them
to be tunable while leaving others 
as default.
For TSP-Single and TSP-Multi
we used a same instance set.
Specifically, we used the $portgen$ and the
$portcgen$ generators from the 8th DIMACS 
Implementation Challenge to generate
1000 ``uniform'' instances 
(in which the cities are randomly distributed) 
and 1000 ``clustering'' instances
(in which the cities are distributed 
around different central points).
The problem sizes of all these generated instances
are within $[1500,2500]$. Once again, we divided them
evenly for training and testing.
The base solver used in TSP-Single was 
LKH version 2.0.7 \cite{helsgaun2000effective} 
(with 35 parameters),
one of the state-of-the-art inexact solver for TSP.
In TSP-Multi, in addition to LKH, we included 
another two powerful TSP solvers, 
GA-EAX version 1.0 \cite{nagata2013powerful} 
(with 2 parameters)
and CLK \cite{applegate2003chained} 
(with 4 parameters),
as the base solvers, resulting in a configuration space 
containing 43 parameters 
(including a top-level parameter used to select a base solver). 

\begin{table}[t]
\footnotesize
\centering
\caption{Summary of the used instance sets, 
cutoff time, base parameterized 
solvers and configuration space size in each scenario.}
\scalebox{0.75}{
\begin{tabular}{p{1.5cm}p{3.5cm}p{1.6cm}p{2.55cm}}
\hline
                 & Instance Set & Cutoff Time & Base Solvers                  \\
\hline
      SAT-Single & From the SAT'16 Competition Agile Track, 2000 instances (1000 for training, 1000 for testing) & 60s  & Riss6 \cite{manthey2016riss}, $|C|=135$\\
      SAT-Multi  & From the SAT'12 Challenge Application Track, 600 instances (300 for training, 300 for testing) & 180s(900s) & 8 solvers considered by \cite{wotzlaw2012pfoliouzk}, $|C|=150$ \\
      TSP-Single & Same as TSP-Multi & 20s & LKH \cite{helsgaun2000effective}, $|C|=35$\\
      TSP-Multi  & 2000 instances containing 1000 ``uniform'' ones and 1000 ``clustering'' ones generated using the generators from the DIMACS TSP Challenge (1000 for training, 1000 for testing) & 20s & LKH \cite{helsgaun2000effective}, CLK \cite{applegate2003chained} and GA-EAX \cite{nagata2013powerful}, $|C|=42$\\
\hline    
\end{tabular}}
\label{tab:scenario}%
\end{table}%

\begin{table}
\footnotesize
\centering
\caption{Detailed time budget (in hours) for each method in each scenario.
In the experiments $r_{pc}$ (for PCIT) 
and $r_{ac}$ (for GLOBAL, $\mathrm{PARHYDRA_{b}}$ and CLUSTERING) 
were both set to 10.
The 3-tuple in each cell represents 
(configuration time budget, validation time budget, total CPU time).
Given the same configuration budget, the same validation budget and $r_{pc}=r_{ac}$,
PCIT, GLOBAL and CLUSTERING would consume the same amount of 
CPU time (See Section~\ref{runtime}). \
Thus M\_group is used to represent these methods for brevity. For $\mathrm{PARHYDRA_{b}}$, the configuration budget   was set to grow linearly with $b$, same as \protect\cite{lindauer2017automatic}.}

\scalebox{0.85}{
\begin{tabular}{p{1.7cm}p{1.45cm}p{1.5cm}p{1.5cm}p{1.45cm}}
\hline
                 & SAT-Single & SAT-Multi & TSP-Single & TSP-Multi \\
\hline
      M\_group & (36,4,3200) & (80,4,6720) & (16,2,1440) & (24,2,2080) \\
      $\mathrm{PARHYDRA}$ & (6,4,3600) & (15,4,6840) & (3,2,1800) & (4,2,2160) \\
      $\mathrm{PARHYDRA_{2}}$ & (12,4,3200) & (30,4,6800) & (6,2,1600) & (8,2,2000) \\
      $\mathrm{PARHYDRA_{4}}$ & (24,4,3360) & (60,4,7680) & (12,2,1680) & (16,2,2160) \\
\hline    
\end{tabular}}
\label{tab:cputime}%
\end{table}

\subsubsection{Competitors and Time Budgets}
Besides PCIT, we implemented GLOBAL, $\mathrm{PARHYDRA_{b}}$ (with b=1,2,4),
and CLUSTERING (with normalization options 
including linear normalization, standard normalization and no normalization), 
as described in \cite{lindauer2017automatic} 
for comparison.
For all considered ACPP methods here, 
SMAC version 2.10.03 \cite{hutter2011sequential}
was used as the algorithm configurator.
Since the performance of SMAC could be often 
improved when used with the instance features,
we gave SMAC access to 
the 126 SAT features 
used in \cite{hutter2011sequential},
and the 114 TSP features used in \cite{kotthoff2015improving}.
The same features were also 
used by PCIT (for transferring instances) 
and CLUSTERING 
(for clustering instances).
To make the comparisons fair, 
the total CPU time consumed by each method 
was kept almost the same.
The detailed setting of time budget for each method 
is given in Table~\ref{tab:cputime}.
To validate whether the instance transfer in PCIT 
is useful, we included another method, named PCRS 
(parallel configuration with random splitting), 
in the comparison.
PCRS differs from PCIT in that it directly configures 
the final portfolios on the initial random instance grouping
and involves no instance transfer.
The time budgets for PCRS were the same as PCIT.

\begin{table*}[t]
\centering
\caption{Results on the test set in the four scenarios. The name of the ACPP method
is used to denote the portfolios constructed by it.
The performance of a solver is shown in boldface if it was not significantly different from the best performance (according to a permutation test with 100000 permutations and significance level $p=0.05$). For CLUSTERING and $\mathrm{PARHYDRA_{b}}$,
the best performance
achieved by their different implementations is reported and the corresponding implementation option,
i.e., the choice of $b$ for $\mathrm{PARHYDRA_{b}}$ and the normalization strategy (``None'' for no normalization, ``Linear'' for linear normalization and ``Standard'' for standard normalization) for CLUSTERING,
is also reported.}
\scalebox{0.8}{
\begin{tabular}{l|p{1.5cm}ll|p{1.5cm}ll|p{1.5cm}ll|p{1.5cm}ll}
      \hline
      \multirow{2}[0]{*}{} & \multicolumn{3}{c}{SAT-Single} & \multicolumn{3}{c}{SAT-Multi} & \multicolumn{3}{c}{TSP-Single} & \multicolumn{3}{c}{TSP-Multi} \\
          & \multicolumn{1}{l}{\#TOS} & \multicolumn{1}{l}{PAR-10} & \multicolumn{1}{l}{PAR-1} & \multicolumn{1}{l}{\#TOS} & \multicolumn{1}{l}{PAR-10} & \multicolumn{1}{l}{PAR-1} & \multicolumn{1}{l}{\#TOS} & \multicolumn{1}{l}{PAR-10} & \multicolumn{1}{l}{PAR-1} & \multicolumn{1}{l}{\#TOS} & \multicolumn{1}{l}{PAR-10} & \multicolumn{1}{l}{PAR-1} \\
      \hline
  Baseline   & 383   & 238   & 31  & 71    & 2275   & 358  & 565  & 118  & 16   & 455    & 99  & 17 \\
    PCRS     & 234   & 152   & 26  & 44    & 1435   & 247  & 110  & 31  & 11   & 105  & 30  & 11\\
    PCIT     & \textbf{181}   & \textbf{119}   & \textbf{21}  & \textbf{35}    & \textbf{1164}   & \textbf{219}  & \textbf{87}   & \textbf{24}  & \textbf{8}    & \textbf{86}   & \textbf{24}  & \textbf{9}\\
    GLOBAL   & 230   & 149   & 25  & 46    & 1495   & 253  & 224  & 53  & 13   & 150  & 41  & 14\\
    $\mathrm{PARHYDRA_{b}}$ & 235 \footnotesize{b=4}  & 151   & 24  & 40 \footnotesize{b=1}   & 1326   & 246  & 107 \footnotesize{b=1}  & 29  & 10   & \textbf{85} \footnotesize{b=2}  & \textbf{24}  & \textbf{9}\\
  CLUSTERING & 227 \footnotesize{None}   & 146     &  23  & 43 \footnotesize{None}     & 1415      & 254    & 121 \footnotesize{Linear}   & 31   & 9   & 99 \footnotesize{Linear}   & 28   & 10 \\
    \hline
\end{tabular}}
\label{tab:results_1}
\end{table*}

\subsubsection{Baselines} 
For each scenario, we identified a sequential solver as  
the baseline by using SMAC to
configure on the training set and the configuration 
space of the scenario.  

  
\subsubsection{Experimental Environment}
All the experiments were conducted on a cluster
of 5 Intel Xeon machines with 60 GB RAM and 6 cores each 
(2.20 GHz, 15 MB Cache), 
running Centos 7.5.

\subsection{Results}
We tested each solver (including the ACPP portfolios and the baseline) 
by running it on each test instance for 3 times,
and reported the median performance.
The obtained number of timeouts (\#TOS), PAR-10 and PAR-1 
are presented in Table~\ref{tab:results_1}.
For CLUSTERING and $\mathrm{PARHYDRA_{b}}$,
we always reported the best performance
achieved by their different implementations.
To determine whether the performance differences between these
solvers were significant, we performed a permutation test (with 100000 permutations and significance level $p=0.05$) to the (0/1) timeout scores, the PAR-10 scores and the PAR-1 scores.
Overall the portfolios constructed by PCIT 
achieved the best performances in Table~\ref{tab:results_1}.
In SAT-Single, SAT-Multi and TSP-Single, it achieved 
significantly and substantially better performances 
than all the other solvers.
Although in TSP-Multi, the portfolio constructed by
$\mathrm{PARHYDRA_{b}}$ obtained slightly better results
than the one constructed by PCIT 
(however the performance difference is insignificant), as aforementioned,
the appropriate value of $b$ varied across different 
scenarios (as shown in Table~\ref{tab:results_1}) 
and for a specific scenario it was actually 
unknown in advance (in TSP-Multi it was 2).
Similarly, as shown in Table~\ref{tab:results_1},
the best normalization strategy for CLUSTERING also
varied across different scenarios.   
Compared to the portfolios constructed by PCRS,
the ones constructed by PCIT 
consistently obtain much better
results, which verified the effectiveness of 
the instance transfer mechanism of PCIT.
Finally, all the ACPP methods here 
could build portfolios that obtained
much better results than the baselines, 
indicating the great benefit by combining complementary configurations obtained from a rich design space.  

To further evaluate the portfolios constructed by PCIT, 
we compared them 
with the state-of-the-art manually designed parallel solvers.
Specifically, we considered the ones constructed for SAT.
We chose Priss6 \cite{manthey2016riss} to compare with 
the one constructed in SAT-Single, 
since Priss6 is the official parallel version of Riss6
(the base solver in SAT-Single).
For the same reason, we chose PfolioUZK \cite{wotzlaw2012pfoliouzk} 
(the gold medal winning solver of the parallel track of the SAT'12 Challenge) 
to compare with the one constructed in SAT-Multi.
Finally, we chose Plingeling (version bbc) \cite{biere2016splatz}, 
the gold medal winning solver of the parallel track of the SAT'16 Competition,
to compare with both.
Note that all the manually designed solvers 
considered here have implemented 
far more advanced solving strategies (e.g., clause sharing) 
than only independently running component solvers in parallel.
In the experiments the default settings of these solvers were used
and the same statistical tests as before were conducted.
As shown in Table~\ref{tab:results_2}, 
on SAT-Single test set, 
the portfolio constructed by PCIT achieved 
much better results than others. 
This may be because the parallel solvers 
considered here are not  
designed for the type 
of these instances (obtained from the SAT'16 Competition Agile track, which is for 
simple fast SAT solvers with low overhead),
which on the other hand 
demonstrates the wide applicability of ACPP methods.
It is impressive that, on SAT-Multi test set, 
the portfolio constructed by
PCIT (regardless of its simple solving strategy) 
obtained slightly better results 
than pfolioUZK, and 
could reach the performance level of the more state-of-the-art Plingeling.
Such results imply that the portfolios constructed by PCIT may be a good staring point for designing more powerful parallel solvers.

\begin{table}
\centering
\caption{Test results of parallel solvers on the test set of SAT-Single and SAT-Multi. The performance of a solver is shown in boldface if it was not significantly different from the best performance (according to a permutation test with 100000 permutations and significance level $p=0.05$).}
\scalebox{0.75}{
\begin{tabular}{l|lll|lll}
      \hline
      \multirow{2}[0]{*}{} & \multicolumn{3}{c}{SAT-Single} & \multicolumn{3}{c}{SAT-Multi} \\
          & \multicolumn{1}{l}{\#TOS} & \multicolumn{1}{l}{PAR-10} & \multicolumn{1}{l}{PAR-1} & \multicolumn{1}{l}{\#TOS} & \multicolumn{1}{l}{PAR-10} & \multicolumn{1}{l}{PAR-1} \\
      \hline
    PCIT   & \textbf{181}   & \textbf{119}   & \textbf{21}   & \textbf{35}  & \textbf{1164}    & \textbf{219}  \\
    Priss6     & 225   & 146   & 25  & -    & -  & -  \\
    PfolioUZK     & -   & -   & -  & \textbf{36}    & \textbf{1185}   & \textbf{213}  \\
    Plinegling-bbc   & 452   & 276   & 32  & \textbf{33}   & \textbf{1090} & \textbf{199}  \\
    \hline
\end{tabular}}
\label{tab:results_2}
\end{table}

\section{Conclusion}
In this paper we proposed a 
novel ACPP method which utilized an 
instance transfer mechanism to improve the quality 
of the instance grouping. The experimental results
verified the effectiveness of the proposed method. 
Directions of future work 
include extending the proposed method 
to use parallel solvers as base solvers, and investigating solving ACPP
from the perspective of subset selection.

\bibliographystyle{named}
\bibliography{ijcai18}

\end{document}